\relax
\documentclass[letterpaper]{article} 
\usepackage{aaai21}  
\usepackage{times}  
\usepackage{helvet} 
\usepackage{courier}  
\usepackage[hyphens]{url}  
\usepackage{graphicx} 
\usepackage{natbib}  
\usepackage{caption} 

\usepackage{amssymb}
\usepackage{amsmath}
\usepackage{mathtools}
\usepackage{multirow}
\usepackage{booktabs}
\usepackage{xcolor}

\usepackage[switch]{lineno}

\usepackage[ruled,linesnumbered]{algorithm2e}

\usepackage{tablefootnote}
\usepackage{threeparttable}

\title{PTN: A Poisson Transfer Network for Semi-supervised Few-shot Learning}
\author{
    Huaxi Huang\textsuperscript{\rm 1}, Junjie Zhang\textsuperscript{\rm 2}, Jian Zhang\textsuperscript{\rm 1}, Qiang Wu\textsuperscript{\rm 1}, Chang Xu \textsuperscript{\rm 3}
    \\
}
\affiliations{
    \textsuperscript{\rm 1}University of Technology Sydney, Sydney NSW 2007, Australia\\
    \textsuperscript{\rm 2}Shanghai University, Shanghai, China\\
    \textsuperscript{\rm 3}The University of Sydney, Sydney NSW 2006, Australia\\
}

\begin{document}
\maketitle

\begin{abstract}

The predicament in semi-supervised few-shot learning (SSFSL) is to maximize the value of the extra unlabeled data to boost the few-shot learner. In this paper, we propose a Poisson Transfer Network (PTN) to mine the unlabeled information for SSFSL from two aspects. 
First, the Poisson Merriman–Bence–Osher (MBO)
model builds a bridge for the communications between labeled and unlabeled examples. This model serves as a more stable and informative classifier than traditional graph-based SSFSL methods in the message-passing process of the labels.
Second, the extra unlabeled samples are employed to transfer the knowledge from base classes to novel classes through contrastive learning. 
Specifically, we force the augmented positive pairs close while push the negative ones distant. Our 
contrastive transfer scheme implicitly learns the novel-class embeddings to alleviate the over-fitting problem on the few labeled data.
Thus, we can mitigate the degeneration of embedding generality in novel classes. 
Extensive experiments indicate that PTN outperforms the state-of-the-art few-shot and SSFSL models on \textit{miniImageNet} and \textit{tieredImageNet} benchmark datasets. 

\end{abstract}

\section{Introduction}
\noindent
Few-shot learning \cite{miller2000learning,fei2006one,vinyals2016matching} aims to learn a model that generalizes well with a few instances of each novel class.
In general, a few-shot learner is firstly trained on a substantial annotated dataset, also noted as the base-class set, and then adapted to unseen novel classes with a few labeled instances.
During the evaluation, a set of few-shot tasks are fed to the learner, where each task consists of a few support (labeled) samples and a certain number of query (unlabeled) data.
This research topic has been proved immensely appealing in the past few years, as a large number of few-shot learning methods are proposed from various perspectives. Mainstream methods can be roughly grouped into two categories. The first one is learning from episodes \cite{vinyals2016matching}, also known as meta-learning, which adopts the base-class data to create a set of episodes. Each episode is a few-shot learning task, with support and query samples that simulate the evaluation procedure.
The second type is the transfer-learning based method, which focuses on learning a decent classifier by transferring the domain knowledge from a model pre-trained on the large base-class set~\cite{chen2018closer,qiao2018few}. This paradigm decouples the few-shot learning progress into representation learning and classification, and has shown favorable performance against meta-learning methods in recent works \cite{tian2020rethinking,ziko2020laplacian}. Our method shares somewhat similar motivation with transfer-learning based methods and proposes to utilize the extra unlabeled novel-class data and a pre-trained embedding to tackle the few-shot problem.

Compared with collecting labeled novel-class data, it is much easier to obtain abundant unlabeled data from these classes. Therefore, semi-supervised few-shot learning (SSFSL)~\cite{ren2018meta,liu2018learning,li2019learning,yu2020transmatch} is proposed to combine the auxiliary information from labeled base-class data and extra unlabeled novel-class data to enhance the performance of few-shot learners.
The core challenge in SSFSL is how to fully explore the auxiliary information from these unlabeled.
Previous SSFSL works indicate that graph-based models~\cite{liu2018learning,ziko2020laplacian} can learn a better classifier than inductive ones~\cite{ren2018meta,li2019learning,yu2020transmatch}, since these methods directly model the relationship between the labeled and unlabeled samples during the inference.
However, current graph-based models adopt the Laplace learning~\cite{zhu2003semi} to conduct label propagation, the solutions of Laplace learning develop localized spikes near the labeled samples but are almost constant far from the labeled samples, \textit{i.e.,} label values are not propagated well, especially with few labeled samples. Therefore, these models suffer from the underdeveloped message-passing capacity for the labels.
On the other hand, most SSFSL methods adapt the feature embedding pre-trained on base-class data (meta- or transfer- pre-trained) as the novel-class embedding. This may lead to the embedding degeneration problem, as the pre-trained model is designed for the base-class recognition, it tends to learn the embedding that represents only base-class information, and lose information that might be useful outside base classes.

To address the above issues, we propose a novel transfer-learning based SSFSL method, named Poisson Transfer Network (PTN). Specifically, \textbf{\textit{to improve the capacity of graph-based SSFSL models in message passing}}, we propose to revise the Poisson model tailored for few-shot problems by incorporating the query feature calibration and the Poisson MBO model. Poisson learning~\cite{calder2020poisson} has been provably more stable and informative than traditional Laplace learning in low label rate semi-supervised problems. However, directly employing Poisson MBO for SSFSL may suffer from the cross-class bias due to the data distribution drift between the support and query data. Therefore, we improve the Poisson MBO model by explicitly eliminating the cross-class bias before label inference. 
\textbf{\textit{To tackle the novel-class embedding degeneration problem}}, we propose to transfer the pre-trained base-class embedding to the novel-class embedding by adopting unsupervised contrastive training \cite{he2020momentum,chen2020simple} on the extra unlabeled novel-class data. Constraining the distances between the augmented positive pairs, while pushing the negative ones distant, the proposed transfer scheme captures the novel-class distribution implicitly. This strategy effectively avoids the possible overfitting of retraining feature embedding on the few labeled instances.

By integrating the Poisson learning and the novel-class specific embedding, the proposed PTN model can fully explore the auxiliary information of extra unlabeled data for SSFSL tasks.
The contributions are summarized as follows:
\begin{itemize}
\item We propose a Poisson learning based model to improve the capacity of mining the relations between the labeled and unlabeled data for graph-based SSFSL.

\item We propose to adapt unsupervised contrastive learning in the representation learning with extra unlabeled data to improve the generality of the pre-trained base-class embedding for novel-class recognition. 

\item 
Extensive experiments are conducted on two benchmark datasets to investigate the effectiveness of PTN, and PTN achieves state-of-the-art performance.
\end{itemize}
\section{Related Work}
\subsection{Few-Shot Learning} 
As a representative of the learning methods with limited samples, \textit{e.g.,} weakly supervised learning~\cite{lan2017robust,zhang2018adversarial}, semi-supervised learning~\cite{zhu2003semi,calder2019properly}, 
few-shot learning can be roughly grouped into two categories: meta-learning models and transfer-learning models. Meta-learning models adopt the episode training mechanism~\cite{vinyals2016matching}, of which metric-based models optimize the transferable embedding of both auxiliary and target data, and queries are identified according to the embedding distances~\cite{sung2018learning,li2019distribution,Simon_2020_CVPR,zhang2020sgone}. Meanwhile, meta-optimization models~\cite{finn2017model,rusu2018meta} target at designing optimization-centered algorithms to adapt the knowledge from meta-training to meta-testing. 
Instead of separating base classes into a set of few-shot tasks, transfer-learning methods~\cite{qiao2018few,gidaris2018dynamic,chen2018closer,qi2018low} utilize all base classes to pre-train the few-shot model, which is then adapted to novel-class recognition. 
Most recently, Tian \textit{et al.} \cite{tian2020rethinking} decouple the learning procedure into the base-class embedding pre-training and novel-class classifier learning. By adopting multivariate logistic regression and knowledge distillation, the proposed model outperforms the meta-learning approaches. 
Our proposed method is inspired by the transfer-learning framework, where we adapt this framework to the semi-supervised few-shot learning by exploring both unlabeled novel-class data and base-class data to boost the performance of few-shot tasks.

\subsection{Semi-Supervised Few-shot Learning (SSFSL)}
SSFSL aims to leverage the extra unlabeled novel-class data to improve the few-shot learning. \citeauthor{ren2018meta} \cite{ren2018meta} propose a meta-learning based framework by extending the prototypical network \cite{snell2017prototypical} with unlabeled data to refine class prototypes. LST \cite{li2019learning} re-trains the base model using the unlabeled data with generated pseudo labels. During the evaluation, it dynamically adds the unlabeled sample with high prediction confidence into testing. In \cite{yu2020transmatch}, TransMatch proposes to initialize the novel-class classifier with the pre-trained feature imprinting, and then employs MixMatch \cite{berthelot2019mixmatch} to fine-tune the whole model with both labeled and unlabeled data.
As closely related research to SSFSL, the transductive few-shot approaches~\cite{liu2018learning,kim2019edge,ziko2020laplacian} also attempt to utilize unlabeled data to improve the performance of the few-shot learning. These methods adopt the entire query set as the unlabeled data and perform inference on all query samples together. For instance, TPN \cite{liu2018learning} employs graph-based transductive inference to address the few-shot problem, and a semi-supervised extension model is also presented in their work.

Unlike the above approaches, in this paper, we adopt the transfer-learning framework and propose to fully explore the extra unlabeled information in both classifier learning and embedding learning with different learning strategies. 

\begin{figure*}[t]
\begin{center}
\includegraphics[width=0.85\linewidth]{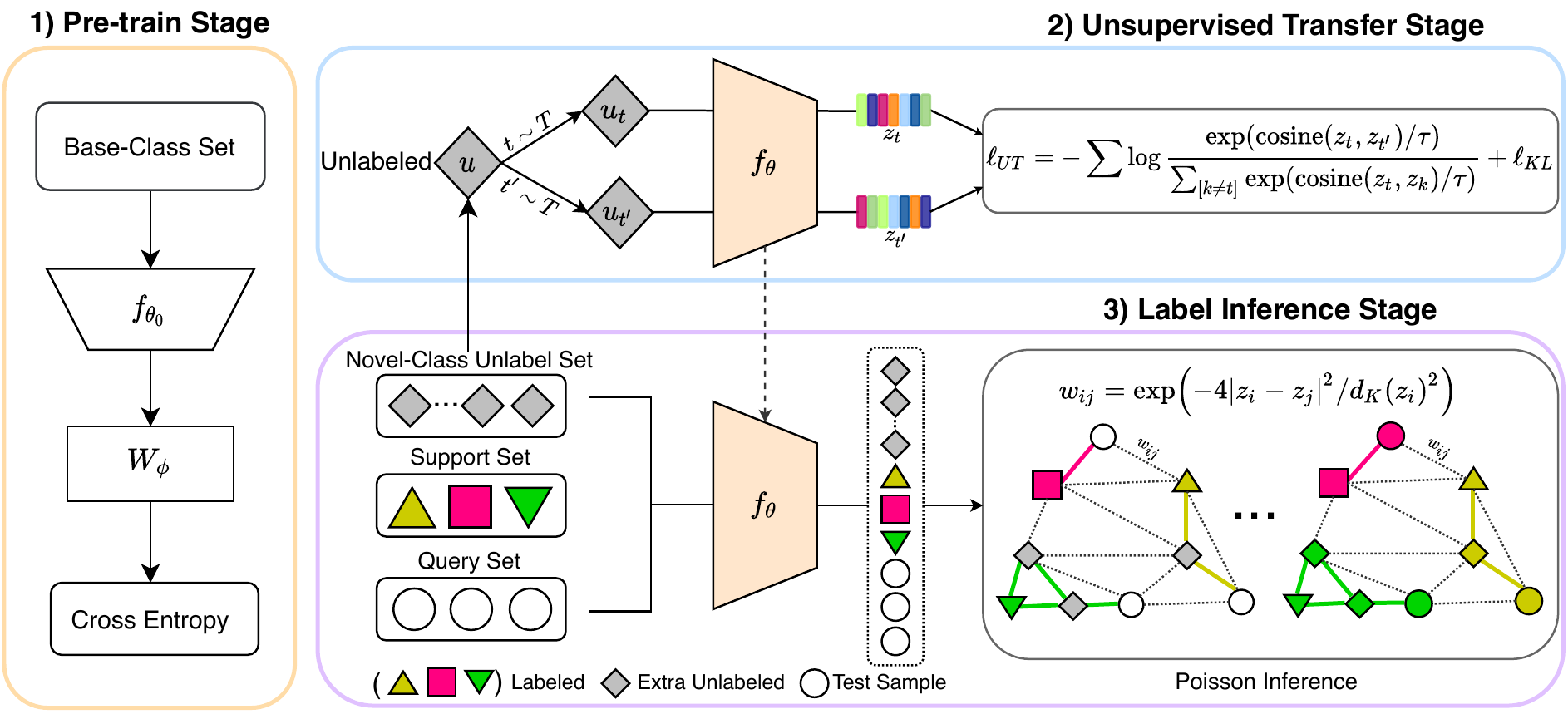} 
\end{center}
   \caption{The overview of the proposed PTN. We first pre-train a feature embedding $f_{\theta_0}$ from the base-class set using standard cross-entropy loss. This embedding is then fine-tuned with the external novel-class unlabeled data by adopting unsupervised transferring loss $\ell_{UT}$ to generate $f_{\theta}$. Finally, we revise a graph model named PoissonMBO to conduct the query label inference.}
\label{framework}
\end{figure*}

\section{Methodology}
\subsection{Problem Definition}\label{def}
In the standard few-shot learning, there exists a labeled support set $S$ of $C$ different classes, ${ S } = \left\{ \left({x} _ { s } ,  {y} _ { s } \right) \right\} _ { s = 1 } ^ { K \times C }$, where $x_s$ is the labeled sample and $y_s$ denote its label. We use the standard basis vector $\mathbf{e}_{i} \in \mathbb{R}^{C}$ represent the $i$-th class, \textit{i.e.}, $y_s \in \left\{\mathbf{e}_{1}, \mathbf{e}_{2}, \ldots, \mathbf{e}_{C}\right\}$. Given an unlabeled query sample $x_q$ from the query set ${ Q } = \left\{  {x} _ { q } \right\} _ { q=1 } ^ { V }$, the goal is to assign the query to one of the $C$ support classes. The labeled support set and unlabeled query set share the same label space, and the novel-class dataset $\mathcal{D}_{novel}$ is thus defined as $\mathcal{D}_{novel} = S \cup Q$. If $S$ contains $K$ labeled samples for each of $C$ categories, the task is noted as a $C$-way-$K$-shot problem.
It is far from obtaining an ideal classifier with the limited annotated $S$. Therefore, few-shot models usually utilize a fully annotated dataset, which has similar data distribution but disjoint label space with $\mathcal{D}_{novel}$ as an auxiliary dataset $\mathcal{D}_{base}$, noted as the base-class set.

For the semi-supervised few-shot learning (SSFSL), we have an extra unlabeled support set ${ U } = \left\{  {x} _ { u } \right\} _ { u = 1 } ^ { N }$. These additional $N$ unlabeled samples are usually from each of the $C$ support classes in standard-setting, or other novel-class under distractor classification settings. Then the new novel-class dataset $\mathcal{D}_{novel}$ is defined as $\mathcal{D}_{novel} = S \cup Q \cup U$. The goal of SSFSL is maximizing the value of the extra unlabeled data to improve the few-shot methods.

For a clear understanding, the details of proposed PTN are introduced as follows: we first introduce the proposed Representation Learning, and then we illustrate the proposed Poisson learning model for label inference.

\subsection{Representation Leaning}
The representation learning aims to learn a well-generalized novel-class embedding through Feature Embedding Pre-training and Unsupervised Embedding Transfer. 
\subsubsection{Feature Embedding Pre-training}
On the left side of Figure \ref{framework}, the first part of PTN is the feature embedding pre-training.

By employing the cross-entropy loss between predictions and ground-truth labels in $\mathcal{D}_{base}$, we train the base encoder $f_{\theta_0}$ in a fully-supervised way, which is the same as \cite{chen2018closer,yu2020transmatch,tian2020rethinking}. 
This stage can generate powerful embedding for the downstream few-shot learner.

\subsubsection{Unsupervised Embedding Transfer} \label{UET}
Directly employ the pre-trained base-class embedding for the novel-class  may suffer from the degeneration problem. However, retraining the base-class embedding with the limited labeled instances is easy to lead to overfitting. How can we train a novel-class embedding to represent things beyond labels when our only supervision is the limited labels? Our solution is unsupervised contrastive learning.
Unsupervised learning, especially Contrastive learning~\cite{he2020momentum,chen2020simple}, recently has shown great potential in representation learning for various downstream vision tasks, and most of these works training a model from scratch. 
However, unsupervised pre-trained models perform worse than fully-supervised pre-trained models. 
Unlike previous works, we propose to adopt contrastive learning to retrain the pre-trained embedding with the unlabeled novel data. In this way, we can learn a decent novel-class embedding by integrating the fully-supervised pre-trained scheme with unsupervised contrastive fine-tuning. 

Specifically, for a minibatch of $n$ examples from the unlabeled novel-class subset ${ U_i = \{x_u\}_{u=1}^n }$, randomly sampling two data augmentation operators $t,t'\in {T}$, we can generate a new feature set $Z = \{ Z_t = \{f_{\theta_0} \circ t(x_u)\}_{u=1}^n \} \cup  \{ Z_{t'} =  \{f_{\theta_0} \circ t'(x_u)\}_{u=1}^n \}$, resulting in $n$ pairs of feature points. We treat each feature pair from the same raw data input as the positive pair, and the other $2(n-1)$ feature points as negative samples. Then the contrastive loss for the minibatch is defined as
\begin{equation}
\ell_{cont}=- \sum_{i,j = 1}^{n} \log \frac{\exp \left(\operatorname{cosine}\left({z}_{i}, {z}_{j}\right) / \tau\right)}{\sum_{k \neq i} \exp \left(\operatorname{cosine}\left({z}_{i}, {z}_{k}\right) / \tau\right)},
\label{NCE}
\end{equation}

where $z_i,z_j$ denote a positive feature pair from $Z$, $\tau$ is a temperature parameter, and $\operatorname{cosine}(\cdot)$ represents the consine similarity. Then, we adopt a Kullback-Leibler divergence ($\ell_{KL}$) between two feature subset $Z_t$ and $Z_{t'}$ as the regulation term. Therefore, the final unsupervised embedding transfer loss $\ell_{UT}$ is defined as 
\begin{equation}
\ell_{UT} = \ell_{cont} + \lambda \ell_{KL}(Z_t ~\|~ Z_{t'}).
\label{UT}
\end{equation}
By training the extra unlabeled data with this loss, we can learn a robust novel-class embedding $f_{\theta}$ from $f_{\theta_0}$.

\subsection{Poisson Label Inference}
Previous studies \cite{zhu2003semi,zhou2004learning,zhu2005semi,liu2018learning,ziko2020laplacian} indicate that the graph-based few-shot classifier has shown superior performance against inductive ones. Therefore, we propose constructing the classifier with a graph-based Poisson model, which adopts different optimizing strategy with representation learning. 
Poisson model~\cite{calder2020poisson} has been proved superior over traditional Laplace-based graph models~\cite{zhu2003semi,zhou2004learning} both theoretically and experimentally, especially for the low label rate semi-supervised problem. 
However, directly applying this model to the few-shot task will suffer from a cross-class bias challenge, caused by the data distribution bias between support data (including labeled support and unlabeled support data) and query data. 

Therefore, we revise this powerful model by eliminating the support-query bias as the classifier. We explicitly propose a query feature calibration strategy before the final Poisson label inference. 
It is worth noticing that the proposed graph-based classifier can be directly appended to the pre-trained embedding without adopting the unsupervised embedding transfer training. We dob this baseline model as \textit{Decoupled Poisson Network} (\textit{DPN}).  

\subsubsection{Query Feature Calibration}
The support-query data distribution bias, also referred to as the cross-class bias~\cite{liu2019prototype}, is one of the reasons for the degeneracy of the few-shot learner. In this paper, we propose a simple but effective method to eliminate this distribution bias for Poisson graph inference. For a SSFSL task, we fuse the labeled support set $S$ and the extra unlabeled set $U$ as the final support set $ B = S \cup U$. We denote the normalized embedded support feature set and query feature set as $Z_b = \{z_b\}$ and $Z_q = \{z_q\}$,
the cross-class bias is defined as 
\begin{equation} \
\begin{split}
& \Delta_{\text {cross}}=\mathbb{E}_{z_{b} \sim p_{\mathcal{B}}}\left[z_{b}\right]-\mathbb{E}_{z_{q} \sim p_{\mathcal{Q}}}\left[z_{q}\right] \\
& \quad\quad~ = \frac{1}{|\mathcal{B}|} \sum_{b=1}^{|\mathcal{B}|} {z}_{b}-\frac{1}{|\mathcal{Q}|} \sum_{q=1}^{|\mathcal{Q}|} {z}_{q}.
\end{split}
\label{QFR}
\end{equation}
We then add the bias $\Delta_{cross}$ to query features. To such a degree, support-query bias is somewhat eliminated. After that, a Poisson MBO model is adopted to infer the query label.

\subsubsection{The Poisson Merriman–Bence–Osher Model}
We denote the embedded feature set as $Z_{novel} = Z_b \cup Z_q = \{z_1, z_2, \dots, z_m\}$ ($m=K\times C + N + V)$, where the first $K \times C$ feature points belong to the labeled support set, the last $V$ feature points belong to the query set, and the remaining $N$ points denote the unlabeled support set. We build a graph with the feature points as the vertices, and the edge weight $w_{ij}$ is the similarity between feature 
point $z_i$ and $z_j$, defined as $w_{i j}=\exp \left(-4\left|z_{i}-z_{j}\right|^{2} / d_{K}\left(z_{i}\right)^{2}\right)$, where $d_{K}\left(z_{i}\right)^{2}$ is the distance between $z_i$ and its $K$-th nearest neighbor. We set $w_{ij} \ge 0$ and $w_{ij} = w_{ji}$. Correspondingly, we define the weight matrix as $W=[w_{ij}]$, the degree matrix as $D=\operatorname{diag}([d_i=\sum_{j=1}^{m}w_{ij}])$, and the unnormalized Laplacian as $L = D - W$. 
As the first $K\times C$ feature points have the ground-truth label, we use $\bar y = \frac{1}{K \times C} \sum_{s=1}^{K\times C} y_s $ to denote the average label vector, and we let indicator $\mathbb{I}_{ij} = 1 $ if $i=j$, else $\mathbb{I}_{ij} = 0 $. The goal of this model is to learn a classifier $g: z \rightarrow \mathbb{R}^{C}$.
By solving the Poisson equation:
\begin{equation}
\begin{split}
& L  g\left(z_{i}\right)=\sum_{j=1}^{K\times C}\left(y_{j}-\bar{y}\right) \mathbb{I}_{ij} \quad \text { for } i=1, \ldots, m,  
\end{split}
\label{PO}
\end{equation}
satisfying $\sum_{i=1}^{m} \sum_{k=1}^{m}w_{ik} g\left(z_{i}\right)=0$, we can then result in the label prediction function $g(z_i)=(g_1(z_i),g_2(z_i),\dots,g_C(z_i))$. The predict label $\hat{y_i}$ of vertex $z_i$ is then determined as $\hat{y_i} = {\arg \max_{j \in\{1, \ldots, C\}} }\left\{g_{j}(x_i)\right\}$. Let $G$ denote the set of $ m \times  C $ matrix, which is the prediction label matrix of the all data. We concatenate the support label to form a label matrix $Y = [y_s] \in \mathbb{R}^{C \times (
K\times C)} $. Let $A = [Y - \bar y, \mathbf{0}^{C \times (m-K\times C)}]$ denotes the initial label of all the data, in which all unlabeled data's label is zero.  The query label of Eq. (\ref{PO}) can be determined by:
\begin{equation}
    G^{tp+1} = G^{tp} + D^{-1} ( A^T - LG^{tp}),
    \label{POSO}
\end{equation}
where $G^{tp}$ denotes the predicted labels of all data at the timestamp $tp$. 
We can get a stable classifier $g$ with a certain number of iteration using Eq. (\ref{POSO}). After that, we adopt a graph-cut method to improve the inference performance by incrementally adjusting the classifier's decision boundary. The graph-cut problem is defined as
\begin{equation}
\min_{g: Z\rightarrow H \atop(g)_{z}=o}\left\{ g^T L g -\mu \sum_{i=1}^{K \times  C}\left(y_{i}-\bar{y}\right) \cdot g\left(z_{i}\right)\right\},
\label{POMBO}
\end{equation}
where $H = \{ \mathbf{e}_{1}, \mathbf{e}_{2}, \ldots, \mathbf{e}_{C} \}$ denotes the annotated samples' label set, $(g)_z = \frac{1}{m}\sum_{i=1}^m g(z_i)$ is the fraction of vertices to each of $C$ classes, and $o =[o_1,o_2,\dots,o_C]^T \in \mathbb{R}^{C}$ is the piror knowledge of the class size distribution that $o_i$ is the fraction of data belonging to class $i$. With the constraint $(g)_z = o$, we can encode the prior 
knowledge into the Poisson Model.
$g^T L g = \frac{1}{2} \sum_{i, j=1}^{m} w_{i j}(g(i)-g(j))^{2}$, this term is the graph-cut energy of the classification given by $g=[g(z_1),g(z_2),\dots,g(z_m)]^T$, widely used in semi-supervised graph models~\cite{zhu2003semi,zhu2005semi,zhou2004learning}.

In Eq. (\ref{POMBO}), the solution will get discrete values, which is hard to solve.
To relax this problem, we use the Merriman-Bence-Osher (MBO) scheme~\cite{garcia2014multiclass} by replacing the graph-cut energy with the Ginzburg-Landau approximation:

\begin{equation}
\begin{split}
& \min _{g\in \mathrm{SP}\{Z\rightarrow \mathbb{R}^C\} \atop(g)_{z}=o}\left\{ \mathrm{GL}_{\tau'} (g) -\mu \sum_{i=1}^{K \times  C}\left(y_{i}-\bar{y}\right) \cdot g\left(z_{i}\right)\right\},\\
& \mathrm{GL}_{\tau'}(g)= g^T L g  +\frac{1}{\tau'} \sum_{i=1}^{m} \prod_{j=1}^{C}\left|g\left(z_{i}\right)-\mathbf{e}_{j}\right|^{2}.
\end{split}
\label{GLPOMBO}
\end{equation}

In Eq. (\ref{GLPOMBO}), $\mathrm{SP}\{Z\rightarrow \mathbb{R}^C\}$ represents the space of projections $g: Z\rightarrow \mathbb{R}^C$, which allow the classifier $g$ to take on any real values, instead of the discrete value from $H$ in Eq. (\ref{POMBO}). More importantly, this leads to a more efficiently computation of the Poisson model. The Eq. (\ref{GLPOMBO}) can be efficiently solved with alternates 
gradient decent strategy, as shown in lines 9-20 of Algorithm \ref{algorithm}. 

\begin{algorithm}[t]
\caption{PTN for SSFSL}\label{algorithm}
\SetKwData{Left}{left}\SetKwData{This}{this}\SetKwData{Up}{up}
  \SetKwFunction{Union}{Union}\SetKwFunction{FindCompress}{FindCompress}
  \SetKwInOut{Input}{Input}\SetKwInOut{Output}{Output}
  \SetKwProg{PoissonMBO}{$PoissonMBO$}{}{$G\leftarrow G[m-V:m,:]$;}
  \Input{$\mathcal{D}_{base}$, $\mathcal{D}_{novel}=S\cup U \cup Q$,\\ $o$, $\mu$, $M_{1}, M_{2}, M_{3}$}
  \Output{Query samples' label prediction $G$}

Train a base model $\mathbf{W}_{\phi} \circ f_{\theta_0} (x)$ with all samples and labels from $\mathcal{D}_{base}$;

Apply unsupervised embedding transfer method to fine-tune the $f_{\theta_0}$ with novel unlabeled data $U$ by using $\ell_{UT}$ in Eq. (\ref{UT}), and result in $f_{\theta}$;

Apply $f_{\theta}$ to extract features on $D_{novel}$ as $Z_{novel}$;

Apply query feature calibration using Eq. (\ref{QFR});

Compute $W, D, L, A$ according to $Z_{novel}$, $G \leftarrow  \mathbf{0}^{m \times C}$

\PoissonMBO{}{

Update $G$ uisng Eq. (\ref{POSO}) with given steps

$\mathrm{d}_{mx} \leftarrow 1 / \max _{1 \leq i \leq m} {D}_{i i}$, $G \leftarrow \mu G$

\For{ $i=1$ \KwTo $M_{1} $}{
\For{ $j=1$ \KwTo $M_{2} $}{${G} \leftarrow {G}-\mathrm{d}_{mx}\left({L} {G}-\mu {A}^{T}\right)$}
$r \leftarrow \textbf{ones}(1,C)$\\
\For{ $j=1$ \KwTo $M_3$}{$\hat{o} \leftarrow \frac{1}{n} \mathbf{1}^{T} \mathbf{P r o j}_{H}({G} \cdot \operatorname{diag}({r}))$\\
${r} \leftarrow \max \left(\min \left({r}+{\varphi} \cdot ({o}-\hat{o}), \upsilon_{\alpha}\right), \upsilon_{\sigma }\right)$
}
$G \leftarrow \mathbf{P r o j}_{H}({G} \cdot \operatorname{diag}({r}))$ }
}
\end{algorithm}
\subsection{Proposed Algorithm}
The overall proposed algorithm is summarized in Algorithm \ref{algorithm}. Inputting the base-class set $\mathcal{D}_{base}$, novel-class set $\mathcal{D}_{novel}$, prior classes' distribution $o$, and other parameters, PTN will predict the query samples' label $G \in \mathbb{R}^{V \times C}$. The query label $\hat{y}_q$ is then determined as $\hat{y}_q = \arg \max _{1 \leq j \leq C} G_{qj}$. 
More specifically, once the encoder $f_\theta$ is learned using the base set $\mathcal{D}_{base}$, we employ the proposed unsupervised embedding transfer method in step 2 in Algorithm \ref{algorithm}. After that, we build the graph with the feature set $Z_{novel}$ and compute the related matrices $W, D, L, A$ in step 3-5. In the label inference stage in steps 6-20, we first apply Poisson model to robust propagate the labels in step 7, and then solve the graph-cut problem by using MBO scheme in several steps of gradient-descent to boost the classification performance. The stop condition in step 7 follow the constraint: $\left\|\mathbf{sp}_{tp}-{W} \mathbf{1} /\left(\mathbf{1}^{T} {W} \mathbf{1}\right)\right\|_{\infty} \leq 1 / m$, where $\mathbf{1}$ is a all-ones 
column vector, $ \mathbf{sp}_{tp}={W} {D}^{-1} \mathbf{sp}_{tp-1}$, $\mathbf{sp_{0}}$ is a $m$-column vector with ones in the first $K\times C$ positions and zeros elsewhere. 
Steps 9-19 are aimed to solve the graph-cut problem in Eq. (\ref{GLPOMBO}), 
To solve the problem, we first divide the Eq. (\ref{GLPOMBO}) into $E_1 = g^TLg-\mu \sum_{i=1}^{K \times  C}\left(y_{i}-\bar{y}\right) \cdot g\left(z_{i}\right)$ and $E_2=\frac{1}{\tau'} \sum_{i=1}^{m} \prod_{j=1}^{C}\left|g\left(z_{i}\right)-\mathbf{e}_{j}\right|^{2}$, and then 
employing the gradient decent alternative on these two energy functions. 
Steps 10-12 are used to optimize the $E_1$.
We optimize the $E_2$ in steps 14-17, $\mathbf{Proj}_{H}: \mathbb{R}^{C} \rightarrow H$ is the closet point projection, $r =[r_1,\dots,r_C]^T$ ($r_i > 0$), ${\varphi}$ is the time step, and $\upsilon_{\alpha}, \upsilon_{\sigma }$ are the clipping values, 
By adopting the gradient descent scheme in steps 14-17, the vector $r$ is generated that also satisfies the constraint $(g)_z = o$ in Eq.(\ref{GLPOMBO}). After obtaining the PoissonMBO's solution $G$, the query samples' label prediction matrix is resolved by step 20.  

The main inference complexity of PTN is $\mathcal{O}(M_1 M_2 E)$
, where $E$ is the number of edges in the graph. As a graphed-based model, PTN's inference complexity is heavier than inductive models. However, previous studies \cite{liu2018learning,calder2020poisson} indicate that this complexity is affordable for few-shot tasks since the data scale is not very big. Moreover, we do not claim that our model is the final solution for SSFSL. We aim to design a new method to make full use of the extra unlabeled information. We report inference time comparison experiments in Table~\ref{time}. The average inference time of PTN is 13.68s.

\section{Experiments}
\subsection{Datasets}
We evaluate the proposed PTN on two few-shot benchmark datasets: miniImageNet and tieredImageNet. The miniImageNet dataset \cite{vinyals2016matching} is a subset of the ImageNet, consisting of 100 classes, and each class contains 600 images of size 84$\times$84. We follow the standard split of 64 base, 16 validation , and 20 test classes  \cite{vinyals2016matching,tian2020rethinking}. The tieredImageNet \cite{ren2018meta} is another subset but with 608 classes instead. We follow the standard split of 351 base, 97 validation, and 160 test classes for the experiments \cite{ren2018meta,liu2018learning}. We resize the images from tieredImageNet to 84$\times$84 pixels, and randomly select $C$ classes from the novel class to construct the few-shot task. Within each class, $K$ examples are selected as the labeled data, and $V$ examples from the rest as queries. The extra $N$ unlabeled samples are selected from the $C$ classes or rest novel classes. We set $C=5, K= \{1,5\}, V=15$ and study different sizes of $N$. We run 600 few-shot tasks and report the mean accuracy with the 95\% confidence interval.

\subsection{Implementation Details}
Same as previous works~\cite{rusu2018meta,dhillon2019baseline,liu2019prototype,tian2020rethinking,yu2020transmatch}, we adopt the wide residual network (WRN-28-10) \cite{zagoruyko2016wide} as the backbone of our base model $W_{\phi} \circ f_{\theta_0}$, and we follow the protocals in \cite{tian2020rethinking,yu2020transmatch} fusing the base and validation classes to train the base model from scratch. We set the batch size to 64 with SGD learning rate as 0.05 and weight decay as $5e^{-4}$. We reduce the learning rate by 0.1 after 60 and 80 epochs. The base model is trained for 100 epochs.

\begin{table*}[t]
\centering
\resizebox{2.0\columnwidth}{!}{%
\begin{tabular}{lcccc}
\hline
\multicolumn{1}{c}{\multirow{2}{*}{Methods}}                              & \multirow{2}{*}{Type}  & \multirow{2}{*}{Backbone} & \multicolumn{2}{c}{miniImageNet}                                                    \\ \cline{4-5} 
\multicolumn{1}{c}{}                                                      &                        &                           & 1-shot                                   & 5-shot                                   \\ \hline
Prototypical-Net \cite{snell2017prototypical}            & Metric, Meta           & ConvNet-256               & 49.42$\pm$0.78                           & 68.20$\pm$0.66                           \\
Relation Network \cite{sung2018learning}            & Metric, Meta           & ConvNet-64                & 50.44$\pm$0.82                           & 65.32$\pm$0.70                           \\
TADAM \cite{oreshkin2018tadam}                      & Metric, Meta           & ResNet-12                 & 58.50$\pm$0.30                           & 76.70$\pm$0.30
                          \\
DPGN \cite{yang2020dpgn}                            & Metric, Meta       & ResNet-12                 & 67.77$\pm$0.32          & 84.60$\pm$0.43   
                          \\
RFS \cite{tian2020rethinking}                       & Metric, Transfer       & ResNet-12                 & 64.82$\pm$0.60                           & 82.14$\pm$0.43
                         \\ \hline
MAML \cite{finn2017model}                           & Optimization, Meta     & ConvNet-64                & 48.70$\pm$1.84                           & 63.11$\pm$0.92                           \\
SNAIL \cite{mishra2018simple}                       & Optimization, Meta     & ResNet-12                 & 55.71$\pm$0.99                           & 68.88$\pm$0.92                           \\
LEO \cite{rusu2018meta}                             & Optimization, Meta     & WRN-28-10                 & 61.76$\pm$0.08                           & 77.59$\pm$0.12                           \\
MetaOptNet \cite{lee2019meta}                       & Optimization, Meta     & ResNet-12                 & 64.09$\pm$0.62                           & 80.00$\pm$0.45                           \\ \hline
TPN \cite{liu2018learning}                          & Transductive, Meta     & ConvNet-64                & 55.51$\pm$0.86                           & 69.86$\pm$0.65                           \\
BD-CSPN \cite{liu2019prototype}                     & Transductive, Meta     & WRN-28-10                 & 70.31$\pm$0.93                           & 81.89$\pm$0.60                           \\
Transductive Fine-tuning \cite{dhillon2019baseline} & Transductive, Transfer & WRN-28-10                 & 65.73$\pm$0.68                           & 78.40$\pm$0.52                           \\
LaplacianShot \cite{ziko2020laplacian}              & Transductive, Transfer & DenseNet                  & 75.57$\pm$0.19                           & 84.72$\pm$0.13                           \\ \hline
Masked Soft k-Means \cite{ren2018meta}              & Semi, Meta             & ConvNet-128               & 50.41$\pm$0.31                           & 64.39$\pm$0.24                           \\
TPN-semi \cite{liu2018learning}                     & Semi, Meta             & ConvNet-64                & 52.78$\pm$0.27                           & 66.42$\pm$0.21                           \\
LST \cite{li2019learning}                           & Semi, Meta             & ResNet-12                 & 70.10$\pm$1.90                           & 78.70$\pm$0.80                           \\ \hline
TransMatch \cite{yu2020transmatch}                  & Semi, Transfer         & WRN-28-10                 & 62.93$\pm$1.11                          & 82.24$\pm$0.59                           \\
DPN (Ours)                                                                      & Semi, Transfer         & WRN-28-10                 & \multicolumn{1}{l}{79.67$\pm$1.06}       & \multicolumn{1}{l}{86.30$\pm$0.95}       \\
PTN (Ours)                                                                       & Semi, Transfer         & WRN-28-10                 & \textbf{82.66$\pm$0.97} & \textbf{88.43$\pm$0.67} \\ \hline \hline
\multicolumn{1}{c}{\multirow{2}{*}{Methods}}                              & \multirow{2}{*}{Type}  & \multirow{2}{*}{Backbone} & \multicolumn{2}{c}{tieredImageNet}                                                  \\ \cline{4-5} 
\multicolumn{1}{c}{}                                                      &                        &                           & 1-shot                                   & 5-shot                                   \\ \hline
Prototypical-Net \cite{snell2017prototypical}            & Metric, Meta           & ConvNet-256               & 53.31$\pm$0.89                           & 72.69$\pm$0.74                           \\
Relation Network \cite{sung2018learning}            & Metric, Meta           & ConvNet-64                & 54.48$\pm$0.93                           & 71.32$\pm$0.78                           \\
DPGN \cite{yang2020dpgn}                            & Metric, Meta       & ResNet-12                 & 72.45$\pm$0.51          & 87.24$\pm$0.39   
                          \\
RFS \cite{tian2020rethinking}                       & Metric, Transfer       & ResNet-12                 & 71.52$\pm$0.69                           & 86.03$\pm$0.49                           \\ \hline
MAML \cite{finn2017model}                           & Optimization, Meta     & ConvNet-64                & 51.67$\pm$1.81                           & 70.30$\pm$1.75                           \\
LEO \cite{rusu2018meta}                             & Optimization, Meta     & WRN-28-10                 & 66.33$\pm$0.05                           & 81.44$\pm$0.09                           \\
MetaOptNet \cite{lee2019meta}                       & Optimization, Meta     & ResNet-12                 & 65.81$\pm$0.74                           & 81.75$\pm$0.53                           \\ \hline
TPN \cite{liu2018learning}                          & Transductive, Meta     & ConvNet-64                & 59.91$\pm$0.94                           & 73.30$\pm$0.75                           \\
BD-CSPN \cite{liu2019prototype}                     & Transductive, Meta     & WRN-28-10                 & 78.74$\pm$0.95                           & 86.92$\pm$0.63                           \\
Transductive Fine-tuning \cite{dhillon2019baseline} & Transductive, Transfer & WRN-28-10                 & 73.34$\pm$0.71                           & 85.50$\pm$0.50                           \\
LaplacianShot \cite{ziko2020laplacian}              & Transductive, Transfer & DenseNet                  & 80.30$\pm$0.22                           & 87.93$\pm$0.15                           \\ \hline
Masked Soft k-Means \cite{ren2018meta}              & Semi, Meta             & ConvNet-128               & 52.39$\pm$0.44                           & 69.88$\pm$0.20                           \\
TPN-semi \cite{liu2018learning}                     & Semi, Meta             & ConvNet-64                & 55.74$\pm$0.29                           & 71.01$\pm$0.23                           \\
LST \cite{li2019learning}                           & Semi, Meta             & ResNet-12                 & 77.70$\pm$1.60                           & 85.20$\pm$0.80                           \\ \hline
DPN (Ours)                                                                       & Semi, Transfer         & WRN-28-10                 & \multicolumn{1}{l}{82.18$\pm$1.06}       & \multicolumn{1}{l}{88.02$\pm$0.72}       \\
PTN (Ours)                                                                      & Semi, Transfer         & WRN-28-10                 & \textbf{84.70$\pm$1.14} & \textbf{89.14$\pm$0.71} \\ \hline
\end{tabular} }
\caption{The 5-way, 1-shot and 5-shot classification accuracy (\%) on the two datasets with 95\% confidence interval. Tne best results are in bold. The upper and lower parts of the table show the results on miniImageNet and tieredImageNet, respectively.}
\label{Res1}
\end{table*}

In unsupervised embedding transfer, the data augmentation $T$ is defined same as \cite{lee2019meta,tian2020rethinking}. For fair comparisons against TransMatch~\cite{yu2020transmatch}, we also augment each labeled image 10 times by random transformations and generate the prototypes of each class as labeled samples. We apply SGD optimizer with a momentum of 0.9. The learning rate is initialized as $1e^{-3}$, and the cosine learning rate scheduler is used for 10 epochs. We set the batch size to 80 with $\lambda = 1$ in Eq. (\ref{UT}).
For Poisson inference, we construct the graph by connecting each sample to its $K$-nearest neighbors with Gaussian weights. We set $K = 30$ and the weight matrix $W$ is summarized with $w_{ii} = 0$, which accelerates the convergence of the iteration in Algorithm \ref{algorithm} without change the solution of the Equation \ref{PO}. We set the max $tp = 100$ in step 7 of Algorithm \ref{algorithm} by referring to the stop constraint discussed in the Proposed Algorithm section. 
We set hyper-parameters $\mu = 1.5, M_1 = 20, M_2 = 40$ and $M_3 = 100$ empirically. Moreover, we set ${\varphi} = 10, \upsilon_{\alpha}=0.5, \upsilon_{\sigma}=1.0$. 
\subsection{Experimental Results}
\subsubsection{Comparison with the State-Of-The-Art}
In our experiments, we group the compared methods into five categories, and the experimental results on two datasets are summarized in Table \ref{Res1}. 
With the auxiliary unlabeled data available, our proposed PTN outperforms the metric-based and optimization-based few-shot models by large margins, indicating that the proposed model effectively utilizes the unlabeled information for assisting few-shot recognition.
By integrating the unsupervised embedding transfer and PoissonMBO classifier, PTN achieves superior performance over both transductive and existing SSFSL approaches. Specifically, under the 5-way-1-shot setting, the classification accuracies are 81.57\% vs. 63.02\%~TransMatch \cite{yu2020transmatch}, 84.70\% vs. 80.30\%~LaplacianShot \cite{ziko2020laplacian} on miniImageNet and tieredImageNet, respectively; under the 5-way-5-shot setting, the classification accuracies are 88.43\% vs. 78.70\%~LST \cite{li2019learning}, 89.14\% vs. 81.89\%~BD-CSPN \cite{liu2019prototype} on miniImageNet and tieredImageNet, respectively.
These results demonstrate the superiority of PTN for SSFSL tasks.
\subsubsection{Different Extra Unlabeled Samples}
\begin{table}[]
\centering
\begin{tabular}{@{}cccc@{}}
\toprule
Methods & Num\_U          & 1-shot         & 5-shot         \\ \midrule
~~PTN$^*$    & \multicolumn{1}{c}{0}   & 76.20$\pm$0.82 & 84.25$\pm$0.61 \\
PTN    & \multicolumn{1}{c}{0}   & 77.01$\pm$0.94 & 85.32$\pm$0.68 \\
PTN    & \multicolumn{1}{c}{20}  & 77.20$\pm$0.92 & 85.93$\pm$0.82 \\
PTN    & \multicolumn{1}{c}{50}  & 79.92$\pm$1.06 & 86.09$\pm$0.75 \\
PTN    & \multicolumn{1}{c}{100} & 81.57$\pm$0.94 & 87.17$\pm$0.58 \\
PTN    & \multicolumn{1}{c}{200} & \textbf{82.66$\pm$0.97} & \textbf{88.43$\pm$0.76} \\ \bottomrule
\end{tabular}
\caption{The 5-way, 1-shot and 5-shot classification accuracy (\%) with different number of extra unlabeled samples on miniImageNet. PTN$^*$ denotes that we adopt PTN as the transductive model without fine-tune embedding. Best results are in bold.}
\label{Res2}
\end{table}

We show the results of using different numbers of extra unlabeled instances in Table \ref{Res2}. For Num\_U = 0, PTN$^*$ can be viewed as the transductive model without extra unlabeled data, where we treat query samples as the unlabeled data, and we do not fine-tune the embedding with query labels for fair comparisons. Contrary to PTN$^*$, the proposed PTN model utilize the query samples to fine-tune the embedding when Num\_U=0.
It can be observed that our PTN model achieves better performances with more extra unlabeled samples, which indicates the effectiveness of PTN in mining the unlabeled auxiliary information for the few-shot problem.

\subsubsection{Results with Distractor Classes}
\begin{figure}[t]
\begin{center}
\includegraphics[width=0.95\linewidth]{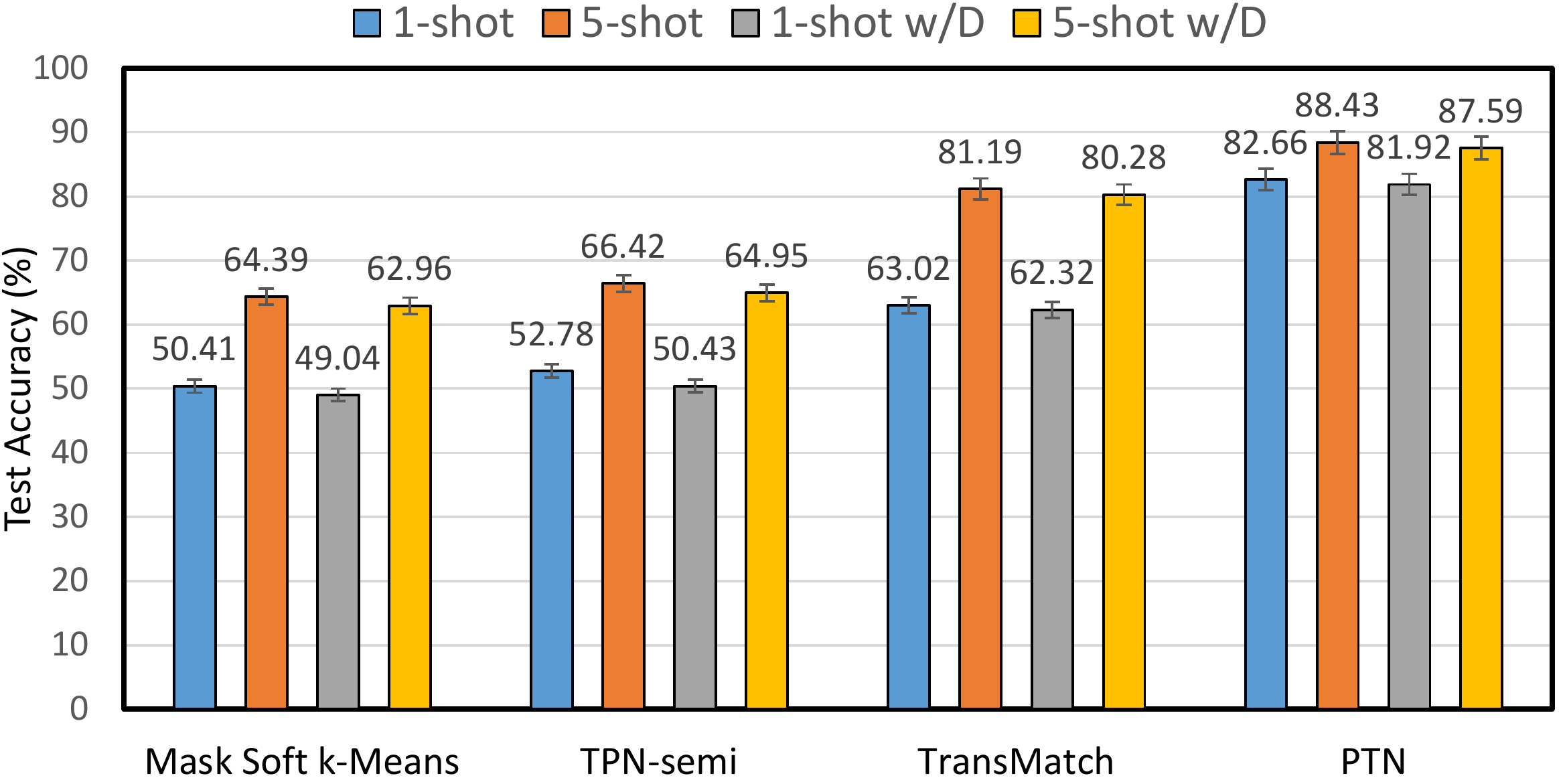} 
\end{center}
\caption{The 5-way, 1-shot and 5-shot classification accuracy (\%) with different number of extra unlabeled samples on miniImageNet. w/D means with distractor classes.}
\label{dist}
\end{figure}

Inspired by \cite{ren2018meta,liu2018learning,yu2020transmatch}, we further investigate the influence of distractor classes, where the extra unlabeled data are collected from classes with no overlaps to labeled support samples. We follow the settings in \cite{ren2018meta,liu2018learning}. As shown in Figure \ref{dist}, even with distractor class data, the proposed PTN still outperforms other SSFSL methods by a large margin, which indicates the robustness of the proposed PTN in dealing with distracted unlabeled data.

\subsection{Ablation Study}
\begin{table}[t]
\begin{minipage}{16.5cm}
\begin{tabular}{@{}lcc@{}}
\toprule
Methods                                            & 1-shot         & 5-shot         \\ \midrule
TransMatch                                         & 62.93$\pm$1.11 & 82.24$\pm$0.59 \\
Label Propagation (LP)                  & 74.04$\pm$1.00 & 82.60$\pm$0.68 \\
PoissonMBO                              & 79.67$\pm$1.02 & 86.30$\pm$0.65 \\
DPN                         & 80.00$\pm$0.83 & 87.17$\pm$0.51 \\
Unsup Trans+LP  \footnote{Unsup Trans means Unsupervised Embedding Transfer.} & 75.65$\pm$1.06 & 84.46$\pm$0.68 \\
Unsup Trans+PoissonMBO        & 80.73$\pm$1.11 & 87.41$\pm$0.63 \\
Unsup Trans+PTN \footnote{PTN consists of Unsup Trans and DPN.}    & \textbf{82.66$\pm$0.97} & \textbf{88.43$\pm$0.76} \\ \bottomrule
\end{tabular}
\end{minipage}
\caption{Ablation studies about the proposed PTN, all methods are based on a pretrained embedding with 200 extra unlabeled samples each class on miniImageNet for 5-way, 1-shot and 5-shot classification (\%). Best results are in bold.}
\label{aby}
\end{table}

We analyze different components of the PTN and summarize the results in Table ~\ref{aby}. All compared approaches are based on the pre-trained WRN-28-10 embedding. 

First of all, we investigate the graph propagation component (classifier).
It can be observed that graph-based models such as Label Propagation~\cite{zhou2004learning} and PoissonMBO~\cite{calder2020poisson} outperform the inductive model TransMatch~\cite{yu2020transmatch}, which is consistent with previous researches~\cite{zhu2005semi,liu2018learning,ziko2020laplacian}. Compared to directly applying PoissonMBO on few-shot tasks, the proposed DPN \textit{\textbf{(without Unsupervised Embedding Transfer)}} achieves better performance, which indicates it is necessary to perform the feature calibration to eliminate the cross-class biases between support and query data distributions before 
label inference.

For investigating the proposed unsupervised embedding transfer in representation learning, we observe that all the graph-based models achieve clear improvement after incorporating the proposed transfer module. For instance, the Label Propagation obtains 1.61\%, 1.86\% performance gains on 5-way-1-shot, and 5-way-5-shot minImageNet classification. These results indicate the effectiveness of the proposed unsupervised embedding transfer. 
Finally, by integrating the unsupervised embedding transfer and graph propagation classifier, the PTN model achieves the best performances compared against all other approaches in Table \ref{aby}. 

\subsection{Inference Time}
We conduct inference time experiments to investigate the computation efficiency of the proposed Poisson Transfer Network (PTN) on the \textit{mini}ImageneNet~\cite{vinyals2016matching} dataset. Same as \cite{ziko2020laplacian}, we compute the average inference time required for each 5-shot task. The results are shown in Table~\ref{time}. Compared with inductive models, the proposed PTN costs more time due to the graph-based Poisson inference. However, our model achieves better classification performance than inductive ones and other transductive models, with affordable inference time.  

\begin{table}[h]
\caption{Average inference time (in seconds) for the 5-shot tasks in \textit{mini}ImageneNet dataset.}
\begin{tabular}{@{}lc@{}}
\toprule
Methods                & Inference Time \\ \midrule
SimpleShot~\cite{wang2019simpleshot}             &      0.009          \\
LaplacianShot~\cite{ziko2020laplacian}          &     0.012         \\
Transductive fine-tune~\cite{dhillon2019baseline} &    20.7            \\
PTN(Ours)              &      13.68          \\ \bottomrule
\end{tabular}

\label{time}
\end{table}

\begin{table*}[t!]
\centering
\caption{Accuracy with various extra unlabeled samples for different semi-supervised few-shot methods on the \textit{mini}ImageNet dataset. All results are averaged over 600 episodes with 95\% confidence intervals. The best results are in bold.}
\begin{tabular}{@{}lccccc@{}}
\toprule
                  & \multicolumn{5}{c}{\textit{mini}ImageNet  5-way-1-shot}                                                        \\ \midrule
                  & 0                       & 20                      & 50                      & 100                     & 200                     \\ \midrule
TransMatch~\cite{yu2020transmatch}        & -                       & 58.43$\pm$0.93          & 61.21$\pm$1.03          & 63.02$\pm$1.07          & 62.93$\pm$1.11          \\
Label Propagation~\cite{zhou2004learning} & 69.74$\pm$0.72& 71.80$\pm$1.02& 72.97$\pm$1.06 & 73.35$\pm$1.05                   & 74.04$\pm$1.00                   \\
PoissonMBO~\cite{calder2020poisson}       & 74.79$\pm$1.06 & 76.01$\pm$0.99 & 76.67$\pm$1.02 & 78.28$\pm$1.02                   &   79.67$\pm$1.02        \\
DPN (Ours)        & 75.85$\pm$0.97 & 76.10$\pm$1.06 & 77.01$\pm$0.92& 79.55$\pm$1.13 & 80.00$\pm$0.83                   \\
PTN (Ours)        & \textbf{77.01$\pm$0.94} & \textbf{77.20$\pm$0.92} & \textbf{79.92$\pm$1.06} & \textbf{81.57$\pm$0.94} & \textbf{82.66$\pm$0.97} \\ \midrule
                  & \multicolumn{5}{c}{\textit{mini}ImageNet  5-way-5-shot}                                                        \\ \midrule
                  & 0                       & 20                      & 50                      & 100                     & 200                     \\ \midrule
TransMatch        & -                       & 76.43$\pm$0.61          & 79.30$\pm$0.59          & 81.19$\pm$0.59          & 82.24$\pm$0.59          \\
Label Propagation & 75.50$\pm$0.60 & 78.47$\pm$0.60 & 80.40$\pm$0.61 & 81.65$\pm$0.59 & 82.60$\pm$0.68                   \\
PoissonMBO        & 83.89$\pm$0.66 & 84.43$\pm$0.67  & 84.94$\pm$0.82 & 85.51$\pm$0.81 & 86.30$\pm$0.65                 \\
DPN (Ours)        & 84.74$\pm$0.63 & 85.04$\pm$0.66 & 85.36$\pm$0.60  & 86.09$\pm$0.63 & 87.17$\pm$0.51                   \\
PTN (Ours)        & \textbf{85.32$\pm$0.68} & \textbf{85.93$\pm$0.82} & \textbf{86.09$\pm$0.75} & \textbf{87.17$\pm$0.58} & \textbf{88.43$\pm$0.76} \\ \bottomrule
\label{unlab}
\end{tabular}
\end{table*}

\subsection{Results with Different Extra Unlabeled}
We conduct further experiments to investigate the current semi-supervised few-shot methods in mining the value of the unlabeled data. All approaches are based on a pre-trained WRN-28-10~\cite{zagoruyko2016wide} model for fair comparisons. As indicated in Table \ref{unlab}, with more unlabeled samples, all the models achieve higher classification performances. However, our proposed PTN model achieves the highest performance among the compared methods, which validates the superior capacity of the proposed model in using the extra unlabeled information for boosting few-shot methods.

\begin{table*}[t]
\centering
\caption{Semi-supervised comparison on the \textit{mini}ImageNet dataset.}
\begin{threeparttable}
\begin{tabular}{@{}lcccc@{}}
\toprule
Methods             & 1-shot         & 5-shot         & \multicolumn{1}{l}{1-shot w/D} & \multicolumn{1}{l}{5-shot w/D} \\ \midrule
Soft K-Means~\cite{ren2018meta} & 50.09$\pm$0.45 & 64.59$\pm$0.28 & 48.70$\pm$0.32 & 63.55$\pm$0.28 \\
Soft K-Means+Cluster~\cite{ren2018meta} & 49.03$\pm$0.24 & 63.08$\pm$0.18 & 48.86$\pm$0.32 & 61.27$\pm$0.24 \\
Masked Soft k-Means~\cite{ren2018meta} & 50.41$\pm$0.31 & 64.39$\pm$0.24 & 49.04$\pm$0.31                 & 62.96$\pm$0.14                 \\
TPN-semi~\cite{liu2018learning}            & 52.78$\pm$0.27 & 66.42$\pm$0.21 & 50.43$\pm$0.84                 & 64.95$\pm$0.73                 \\
TransMatch~\cite{yu2020transmatch}          & 63.02$\pm$1.07 & 81.19$\pm$0.59 & 62.32$\pm$1.04                 & 80.28$\pm$0.62                 \\ \midrule
PTN (Ours)                & \textbf{82.66$\pm$0.97} & \textbf{88.43$\pm$0.67} & \textbf{81.92$\pm$1.02} & \textbf{87.59$\pm$0.61} \\ \bottomrule
\end{tabular}
\begin{tablenotes}
\item[$\divideontimes$]``w/D" means with distraction classification. In this setting, many extra unlabeled samples are from the distraction classes, which is different from the support labeled classes. All results are averaged over 600 episodes with 95\% confidence intervals. The best results are in bold. 
\end{tablenotes}
\end{threeparttable}
\label{mini}
\end{table*}

\begin{table*}[!hpbt]
\centering
\caption{Semi-supervised comparison on the \textit{tiered}ImageNet dataset.}
\begin{threeparttable}
\begin{tabular}{@{}lcccc@{}}
\toprule
Methods             & 1-shot         & 5-shot         & \multicolumn{1}{l}{1-shot w/D} 
& \multicolumn{1}{l}{5-shot w/D} 
\\ \midrule
Soft K-Means~\cite{ren2018meta} & 51.52$\pm$0.36 & 70.25$\pm$0.31 & 49.88$\pm$0.52 & 68.32$\pm$0.22 \\
Soft K-Means+Cluster~\cite{ren2018meta} & 51.85$\pm$0.25 & 69.42$\pm$0.17 & 51.36$\pm$0.31 & 67.56$\pm$0.10 \\
Masked Soft k-Means~\cite{ren2018meta} & 52.39$\pm$0.44 & 69.88$\pm$0.20 & 51.38$\pm$0.38& 69.08$\pm$0.25                 \\
TPN-semi~\cite{liu2018learning}            & 55.74$\pm$0.29 & 71.01$\pm$0.23 & 53.45$\pm$0.93& 69.93$\pm$0.80                 \\ \midrule
PTN (Ours)                & \textbf{84.70$\pm$1.14} & \textbf{89.14$\pm$0.71} & \textbf{83.84$\pm$1.07}                 & \textbf{88.06$\pm$0.62} \\ \bottomrule
\end{tabular}
\begin{tablenotes}
\item[$\divideontimes$]``w/D" means with distraction classification. In this setting, many extra unlabeled samples are from the distraction classes, which is different from the support labeled classes. All results are averaged over 600 episodes with 95\% confidence intervals. The best results are in bold. 
\end{tablenotes}
\end{threeparttable}
\label{tiered}
\end{table*}

\subsection{Results with Distractor Classification}
We report the results of the proposed PTN on both \textit{mini}ImageNet~\cite{vinyals2016matching} and \textit{tiered}ImageneNet~\cite{ren2018meta} datasets under different settings in Table~\ref{mini} and Table~\ref{tiered}, respectively. It can be observed that the classification results of all semi-supervised few-shot models are degraded due to the distractor classes. However, the proposed PTN model still outperforms other semi-supervised few-shot methods with a large margin. This also indicates the superiority of the proposed PTN model in dealing with the semi-supervised few-shot classification tasks over previous approaches.

\section{Conclusion}

We propose a Poisson Transfer Network (PTN) to tackle the semi-supervised few-shot problem, aiming to explore the value of unlabeled novel-class data from two aspects. We propose to employ the Poisson learning model to capture the relations between the few labeled and unlabeled data, which results in a more stable and informative classifier than previous semi-supervised few-shot models. Moreover, we propose to adopt the unsupervised contrastive learning to improve the generality of the embedding on novel classes, which avoids the possible over-fitting problem when training with few labeled samples. 
Integrating the two modules, the proposed PTN can fully explore the unlabeled auxiliary information boosting the performance of few-shot learning.
Extensive experiments indicate that PTN outperforms state-of-the-art few-shot and semi-supervised few-shot methods.

\section*{Acknowledgment}

The authors greatly appreciate the financial support from the Rail Manufacturing Cooperative Research Centre (funded jointly by participating rail organizations and the Australian Federal Government’s Business-Cooperative Research Centres Program) through Project R3.7.3 - Rail infrastructure defect detection through video analytics.

\small
\bibliography{aaai}

\end{document}